\documentclass{article}

\usepackage{arxiv}

\usepackage[utf8]{inputenc} 
\usepackage[T1]{fontenc}    
\usepackage{hyperref}       
\usepackage{url}            
\usepackage{booktabs}       
\usepackage{amsfonts}       
\usepackage{nicefrac}       
\usepackage{microtype}      
\usepackage{cleveref}       
\usepackage{lipsum}         
\usepackage{doi}

\usepackage[dvipdfmx]{graphicx,xcolor}
\graphicspath{{img/}}

\title{Phased Data Augmentation for Training a Likelihood-Based Generative Model \\ with Limited Data}

\author{ \href{}{\hspace{1mm}Yuta Mimura} \\
	Department of Physics\\
	Hokkaido University\\
	Sapporo 060-0810, Japan\\
	\texttt{y-mimura@particle.sci.hokudai.ac.jp} \\
}

\renewcommand{\shorttitle}{Phased Data Augmentation}

\hypersetup{
pdftitle={Phased Data Augmentation for Training a Likelihood-Based Generative Model with Limited Data},
pdfsubject={cs.CV},
pdfauthor={Yuta Mimura},
pdfkeywords={training with limited data, PixelCNNs, VQ-VAE-2, data augmentation, generative models},
}

\begin{document}
\maketitle

\begin{abstract}
Generative models excel in creating realistic images, yet their dependency on extensive datasets for training presents significant challenges, especially in domains where data collection is costly or challenging. 
Current data-efficient methods largely focus on GAN architectures, leaving a gap in training other types of generative models. 
Our study introduces ``phased data augmentation" as a novel technique that addresses this gap by optimizing training in limited data scenarios without altering the inherent data distribution. 
By limiting the augmentation intensity throughout the learning phases, our method enhances the model's ability to learn from limited data, thus maintaining fidelity. 
Applied to a model integrating PixelCNNs with VQ-VAE-2, our approach demonstrates superior performance in both quantitative and qualitative evaluations across diverse datasets. 
This represents an important step forward in the efficient training of likelihood-based models, extending the usefulness of data augmentation techniques beyond just GANs.
\end{abstract}

\keywords{training with limited data, PixelCNNs, VQ-VAE-2, data augmentation, generative models}

\section{Introduction}

Generative models, renowned for their ability to generate compelling images, traditionally rely on large datasets for optimal training. 
However, the challenge of amassing substantial, domain-specific data remains a significant barrier, especially given that 
``it is widely known that data collection is an extremely expensive process in many domains, e.g. medical
images" \cite{tran2021data}. 
Addressing this limitation forms the cornerstone of our study, wherein we introduce a straightforward approach for training generative models efficiently on limited datasets.

In the broader landscape of deep learning, transfer learning has emerged as a potent tool, particularly when large datasets are unavailable. 
While transfer learning can enhance performance in a target domain using data from a related source task, ``when transferring knowledge from a less related source, it may inversely hurt the target performance, a phenomenon known as negative transfer" \cite{wang2019characterizing}.
Concurrently, while data augmentation has been embraced as a strategy to expand datasets artificially, it occasionally distorts the original data distribution.
Furthermore, most existing data-efficient training methods \cite{tran2021data,zhao2020differentiable,karras2020training} primarily cater to the Generative Adversarial Networks (GANs) domain, leaving other generative models relatively unexplored.

Our proposed method, termed ``phased data augmentation", provides the first step in this direction. 
It reduces the intensity of data augmentation in line with the model's learning phases. 
Initially, it amplifies the dataset's effective size, aiding the model in grasping the general patterns. 
As training advances, the augmentation parameters are tightened to ensure the model focuses on salient features intrinsic to the original training data. 
This methodology, based on standard data augmentation, is applicable to generative models other than GANs. 

This paper applies our method to a model that integrates PixelCNNs with VQ-VAE-2, which we refer to as PC-VQ2 \cite{razavi2019generating}. 
PC-VQ2 is a likelihood-based generative model without GANs' architecture. 
PixelCNNs have promising potential to generate images that are both sharper and more varied than many of their counterparts. 
Their integration with VQ-VAE-2 can further enhance image precision, a benefit derived from its hierarchical structure.

Empirical evidence from our experiments consistently demonstrates the superiority of our approach over standard data augmentation technique in both quantitative and qualitative assessments, underscoring its potential as an effective strategy for training a likelihood-based model with limited datasets.  
The robustness of this efficacy, validated across various data domains and sampled datasets, underscores the method's consistent performance improvements over the traditional data augmentation, even in the context of limited data resources. 

The remainder of this paper is organized as follows:
Section 2 situates PC-VQ2 within the broader landscape of generative models, elucidating the merits of PC-VQ2, utilized in our experiments.
In Section 3, we provide an overview of the model under consideration in this study. 
Section 4 delves into our proposed strategy and its application to the aforementioned model. 
The experimental results are detailed in Section 5.
Section 6 discusses related works pertaining to data augmentation. 
This section aims to furnish the reader with a comprehensive background leading to the introduction of phased data augmentation.
Finally, Section 7 offers a concise summary and conclusion of our findings and contributions.

\section{The Position of PC-VQ2 in Generative Models}
Given the expansive realm of generative models, our focus in this paper is on the integration of PixelCNNs with VQ-VAE-2 \cite{razavi2019generating}, underpinned by several compelling attributes.

PixelCNNs have the capability to generate a more diverse set of images compared to prominent generative models, such as GANs \cite{goodfellow2014generative,aggarwal2021generative}. 
This capacity for diversity is not exclusive to PixelCNNs but is also seen in other likelihood-based models.
A notable mention here is Diffusion Models (DMs) \cite{sohl2015deep}, which have recently witnessed a rise in adoption and share this advantageous trait.

What sets PixelCNNs apart, however, is their unique capacity to produce images that are not only diverse but also sharper compared to other likelihood-based models, including DMs.
This edge is ascribed to PixelCNN's pixel-wise learning approach, in stark contrast to the more common image-wise learning seen in other likelihood-based models.
When producing high-fidelity images, this distinction is particularly significant. 

The integration of VQ-VAE-2 further enhances the sharpness, due to its hierarchical structure and discrete latent spaces.
The encoding within these spaces significantly improves the computational efficiency, when PixelCNN models generate pixels for each image in an autoregressive manner.

\section{Overview of the PixelCNNs and VQ-VAE-2 Utileized in PC-VQ2} 
This section provides a comprehensive overview of PixelCNNs and VQ-VAE-2 as utilized in the PC-VQ2 framework. 

VQ-VAE-2 is an AutoEncoder with vector-quantized latent spaces, comprising an encoder $E()$ and decoder $D()$.
When encoding continuous vectors, these are mapped to their nearest quantized ones $\mathbf{e}$:
\begin{equation}
\mbox{Quantize}(E(\mathbf{x})) \!=\! \mathbf{e}_k \; \mbox{where} \; k \!=\! \arg\min_j ||E(\mathbf{x}) \! - \! \mathbf{e}_j||
\end{equation}
\cite{razavi2019generating}.
In the original VQ-VAE paper \cite{van2017neural}, the posterior categorical distribution was considered deterministic, leading to a consistent KL divergence regularization term through a simple uniform prior over z = k.
The model is trained to both maximize the likelihood and tune the matching for quantization.
The loss function is as follows:
\begin{eqnarray}
\mathcal{L}(\mathbf{x}, D(\mathbf{e})) = ||\mathbf{x} \!-\! D(\mathbf{e})||^2_2 \!+\! ||\mbox{sg}[E(\mathbf{x})] \!-\! \mathbf{e}||^2_2  \mbox{}  \!+\! \beta ||\mbox{sg}[\mathbf{e}] \!-\! E(\mathbf{x})||^2_2
\end{eqnarray}
\cite{razavi2019generating}.
Here, $\mbox{sg}$ denotes the ``stop gradient" operation.
The last term encourages the encoder's output to gravitate toward the selected quantized vector.
Additionally, VQ-VAE-2 utilizes a hierarchical structure with two different sizes of quantized latent spaces: the larger ``bottom-level" latent map receives the encoded continuous vectors and the discretized output of the smaller ``top-level" latent map.

PixelCNNs model the joint distribution of pixels in an image $\mathbf{x}$ as the
subsequent product of conditional distributions, where $x_i$
denotes an individual pixel:
\begin{equation}
p(\mathbf{x}) = \prod_{i=1}^{n} p(x_i | x_1, \dots, x_{i-1})
\end{equation}
\cite{van2016conditional}.
The pixel sequence adheres to raster scan ordering.
The dependencies are modeled using masked convolution filters, offering a training efficiency advantage over RNN structures.
The training objective is to maximize the likelihood, with the loss function being a reconstruction loss typified by cross-entropy.

As depicted in Fig.\,1, the PC-VQ2 architecture assigns global image information to the top-level, whereas the bottom-level focuses on local details.
For sampling images, PixelCNNs are employed over the latent maps, as illustrated in Fig.\,A.1 of Appendix A, which provides an overview of the PC-VQ2 sampling process.
\begin{figure*}[t] 
\centering
\begin{center}
\includegraphics[width=15cm]{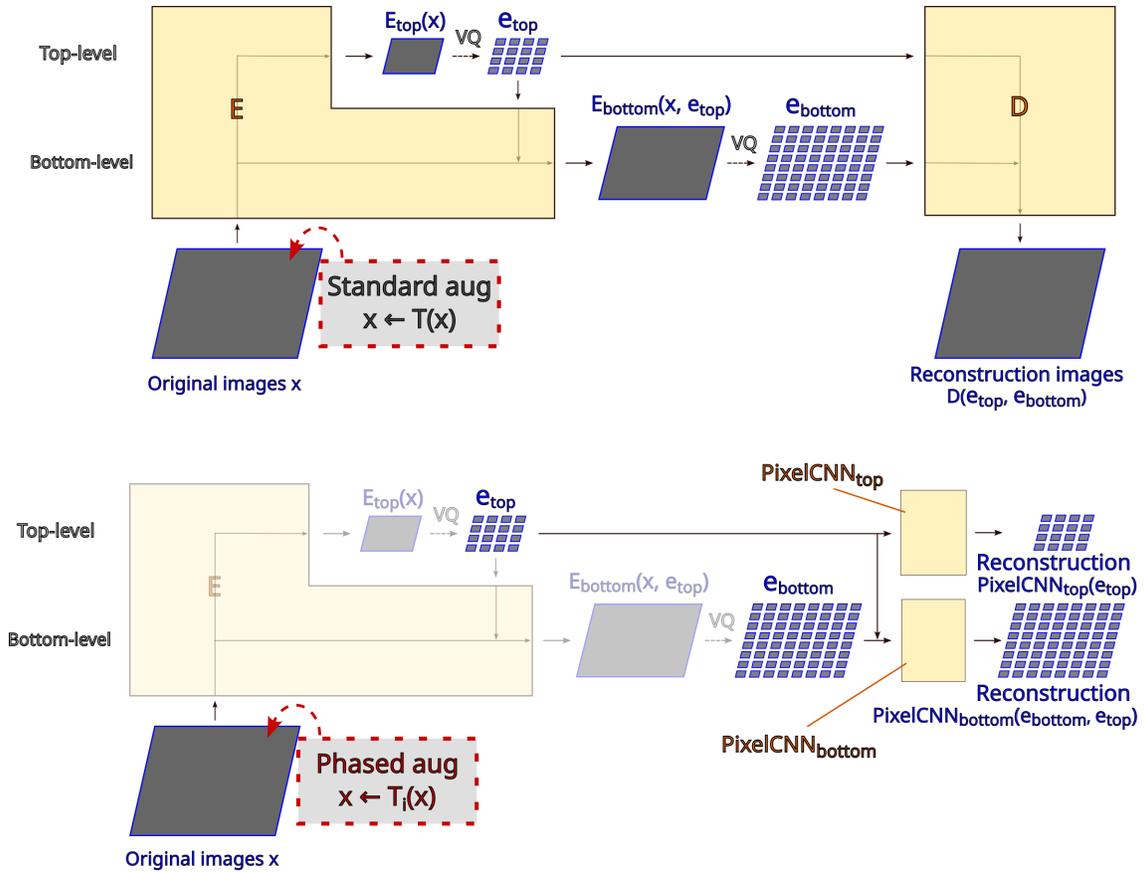}
\end{center}
\label{fig:PC-VQ2_training.eps}
\caption{Overview of a PC-VQ2 training structure and application of phased augmentation to it. The upper figure illustrates the training of VQ-VAE-2, and the lower figure illustrates the individual training of PixelCNNs.}
\end{figure*}

We chose specific PixelCNNs models, that have features mentioned in the VQ-VAE-2 paper, for each latent map.
The bottom-level utilizes a conditional Gated PixelCNN \cite{van2016conditional} featuring gated activation units and conditioning with the top-level.
In contrast, the top-level employs PixelSNAIL \cite{chen2018pixelsnail}, which integrates attention layers while inheriting the gated activation units.
We implemented the VQ-VAE-2 encoder-decoder model, the conditional Gated PixelCNN model, and the PixelSNAIL model on the basis of the original papers \cite{razavi2019generating,van2016conditional,chen2018pixelsnail} and open-source implementations.

Training proceeds initially with VQ-VAE-2, followed by individual training for the top-level PixelSNAIL and the bottom-level Gated PixelCNN.

\section{Phased Data Augmentation}
In this section, we introduce our proposed method, termed ``phased data augmentation", and further explain its applications within PC-VQ2. 

\subsection{Overfitting in Generative Models} 
Generative models often necessitate extensive training datasets, commonly spanning tens of thousands of images, to ensure optimal performance. 
When these models have access only to a limited dataset, overfitting becomes a significant concern.

For instance, when GANs are trained on a constrained dataset using standard or no augmentation, they typically struggle to produce high-quality images \cite{tran2021data,zhao2020differentiable,karras2020training}.

A similar issue arises with the PC-VQ2 model.
When PC-VQ2 is trained with limited data and no data augmentation, it predominantly generates images that resemble noise. 
Even with the incorporation of standard data augmentation techniques, the PC-VQ2 often produces images that appear unnaturally distorted. 
Although standard data augmentation can mitigate overfitting to some extent and enhance the overall quality of the generated images, the output still often appears unnatural.
A detailed comparison of these phenomena is provided in the ``Results and Discussions" Section, specifically in Fig.\,4 and 5.
\begin{figure*}[tb] 
\centering
\begin{center}
\includegraphics[width=15cm]{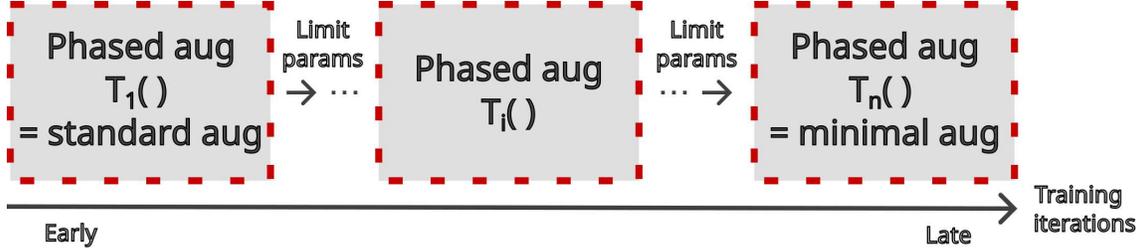}
\end{center}
\label{fig:Phased_aug.eps}
\caption{Graphical representation of phased data augmentation.}
\end{figure*}

\subsection{Methodology of Phased Data Augmentation} 

To address the issue, we introduce a novel yet straightforward training strategy, termed ``phased data augmentation".
The fundamental principle of this approach is described as follows:
\begin{enumerate}
    \item Begin with the full range of label-preserving standard data augmentations on a limited dataset. 
    \item As training progresses, restrict these ranges in phases. 
    \item Shift from standard augmentation to minimal augmentation that maintains a target distribution of the training set.
\end{enumerate}
Figure 2 provides a graphical representation of this strategy.

Because our phased data augmentation builds upon the standard data augmentation technique, it does not require GANs' structures, offering an approach that is applicable across a wide range of model architectures beyond just GANs, by introducing our phased data augmentation as a replacement for standard data augmentation technique.

In this paper, we used basic transformations such as flipping, rotation, zooming which consisted of anisotropic random integer upscaling and constant range cropping, and color-space transformations such as brightness, saturation, and contrast manipulation.
Flipping was consistently used throughout all phases because it tends to maintain a target training set distribution.
The remaining three transformations, rotation, zooming, and color-space transformations, were gradually limited in phases in the order mentioned, to bring the dataset distribution incrementally close to its original state.
We determined the sequence of imposing limitations on the operations by considering their impact on altering the distribution of the original training data. 
Specifically, because rotation, zooming, and color-space transformations progressively exert a stronger influence on the data's distribution, we imposed limitations in this specific order.

A distinct point of this strategy over transfer learning is its capacity to train on a dataset with a distribution that mirrors the original.
This can be seen as an advantage.

As highlighted earlier, during the initial training stages, data augmentation enables a model to grasp the overarching patterns shared by augmented training data from scratch.
As training advances and augmentation is reduced, the model becomes finely attuned to the distinct attributes of the original dataset.
This observation is supported by Fig.\,3, showcasing images generated at each phase in a PC-VQ2 training.
\begin{figure*}[t] 
\centering
\begin{center}
\includegraphics[width=15cm]{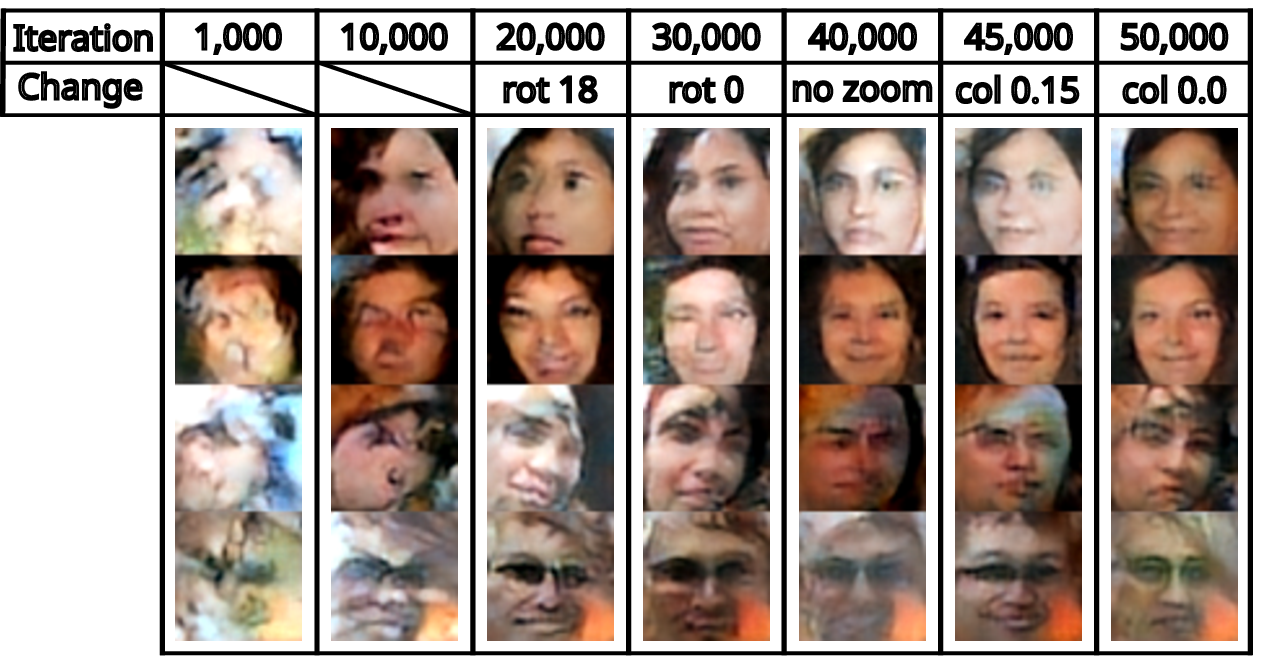}
\end{center}
\label{fig:PC_VQ2_Face_Process.eps}
\caption{Images reconstructed in each phase through a bottom-level PixelCNN training. Each element in the `Change' row indicates the specific limitation imposed relative to the previous phase, with more detailed explanations provided in Section 4.3.}
\end{figure*}

\subsection{Application of Phased Augmentation to PC-VQ2} 
In this section, we delineate the specific setup of phased data augmentation applied to the PC-VQ2 model, based on the methodology described earlier. 
Although certain decisions were made based on empirical observations, they were, as much as possible, carefully grounded in reasoned judgment.
Importantly, in our experiments, the identical setup was applied across multiple data domains and sample datasets, demonstrating clear and consistent superiority over standard data augmentation approach.
Detailed analyses of these experimental outcomes are further elaborated in the ``Results and Discussion" section.

During the initial phase, the following augmentations were applied:
\begin{itemize}
 \item Flipping
 \item Rotation, allowing for a maximum range of $\pm$180 degrees
 \item Zooming
 \item Color-space transformations with a parameter value of 0.30.
\end{itemize}

As the training progressed through subsequent phases:
\begin{itemize}
 \item In the second phase, the rotation's maximum degree was restricted to $\pm$18.
 \item In the third phase, no rotation was allowed.
 \item The fourth phase excluded the zooming operation.
 \item During the fifth phase, the color transformation parameter was set to 0.15.
 \item In the final phase, no color transformation was applied.
\end{itemize}

Regarding the initial settings, the decision to start with rotations up to 360 degrees aimed to maximize the benefits of data augmentation.
For zooming and color transformations, we exercised caution to avoid extreme transformations that could significantly deviate the data from its original distribution, such as overly zoomed-in images where only a part of the subject is visible or overly brightened images where the subject becomes indiscernible.

Concerning rotation, we implemented it in two stages to suppress the impact on the augmented training data's distribution, initially limiting the rotation in the second phase to one-tenth of the full range, because performing the rotation in one go could significantly alter the training data distribution.
As for color transformations, considering the diminished effect of augmentation on limited data in the last phases, we opted for a two-phase approach to encourage learning by retaining learned content while removing excess color transformed information. 
Considering that reducing the augmentation effect to one-tenth would be excessively restrictive, we chose to reduce it to one-half instead.
To prevent the introduction of margins due to the rotation and zooming operations, constant integer upscaling was utilized.
In the rotation process, after upscaling, a consistent center cropping was used to maintain the original resolution.
Specifically, the zooming process employed anisotropic random integer upscaling, wherein the height and width are independently scaled by a factor ranging between 1.05 and 1.30.
After upscaling, cropping was performed to keep the image within the minimum upscaled pixel region.

In terms of phase progression, each phase transitioned after 10,000 iterations, with the exception of the transition from the fifth to the last phase and the conclusion of the final phase, both taking place after 5,000 iterations, considering the diminished effect of augmentation on limited data in the last phases.

All the random values were sampled from uniform distributions.

The codes were implemented using Tensorflow 2.10.1 and OpenCV 4.8.0.

The phased augmentation approach is utilized on the limited training dataset when training the top-level and bottom-level PixelCNNs models, depicted in Fig.\,1.
For VQ-VAE-2 training, the standard augmentation technique is utilized, given that the augmented dataset gets leveraged during the training of the PixelCNNs models.

\section{Results and Discussions}
This section represents the experimental results comparing phased data augmentation to the other data-efficient technique when training PC-VQ2 with limited data.

For the experiments involving PC-VQ2, standard data augmentation technique was adopted for the comparison. 
This decision was based on the observation that using no data augmentation, the PC-VQ2 generated images of negligible quality, resembling noise. 

The FID score was employed to measure performance, utilizing 5,000 generated images and 100 real images, as used in a prior paper on data-efficient GANs \cite{zhao2020differentiable}.
We utilized the first eight sampled images for figures of the generated images.
Details on our experimental settings, can be found in Appendices B.

We utilized the FFHQ dataset, which is the high-quality human-face dataset \cite{karras2019style}, and the AFHQ v2 cat dataset, which is the high-quality cat-face dataset \cite{choi2020starganv2,parmar2022aliased,Karras2021} during the training.
Three distinct subsets were drawn from the FFHQ dataset, specifically indices 0-99, 10,000-10,099, and 20,000-20,099.
For the AFHQ v2 cat dataset, a set of 100 images was selected at random.

As summarized in Table 1, across all training datasets, phased augmentation demonstrated significantly better FID scores compared to standard augmentation technique. 
The scores for each method were clustered around similar values.
This observation, coupled with the consistent and clear superiority of the proposed method over standard data augmentation across various data domains and sampled datasets, underscores the robustness of its efficacy, even in contexts of limited data resources.

Figure 4 showcases the generated images, corroborating the quantitative findings.
Figure 5 illustrates that the generated images during training with no augmentation resemble noise.

The generated images exhibit some blur.
One of the factors is the employment of low-resolution latent spaces, 16-8.
In fact, leveraging higher-resolution latent spaces, 32-16, results in sharper images, albeit with some distortion, as illustrated in Fig.\,A.2.
However, even this resolution remains below the original study's configuration.
The other factors are elaborated upon in Appendix D.

\begin{table*}[b]
\centering
\caption{FID results over trained PC-VQ2 models.}
\label{FID results over trained PC-VQ2 models}
\begin{tabular}{|c|c|c|c|c|}
\hline
Method & FFHQ 0-99 & 10,000-10,099 & 20,000-20,099 & AFHQ v2 Cat \\
\hline
Standard data augmentation & 263.21 & 240.66 & 252.87 & 259.24  \\
\hline
Phased data augmentation  & \textbf{169.62}  & \textbf{140.46} & \textbf{149.63} & \textbf{177.33}  \\
\hline
\end{tabular}
\end{table*}

\clearpage

\begin{figure*}[t]
\begin{center}
  \includegraphics[width=15cm]{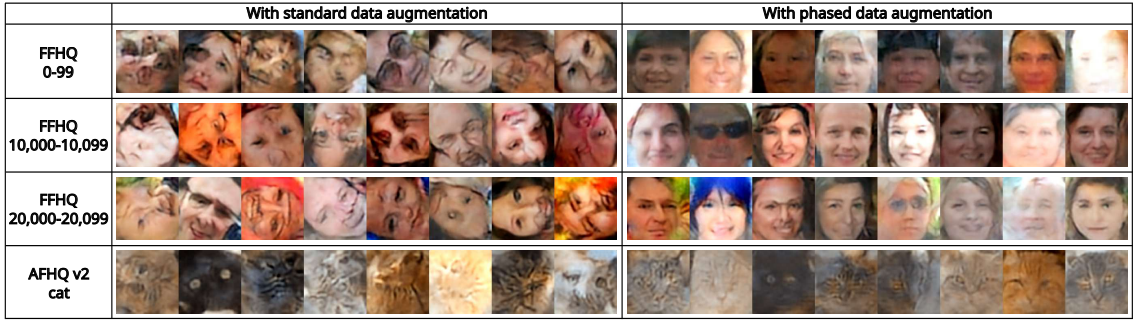}
\end{center}
\label{fig:PC-VQ2_generated_images.eps}
\caption{Generated human-face and cat-face images by the trained PC-VQ2 models.}
\end{figure*}
\begin{figure*}[t]
\begin{center}
  \includegraphics[width=15cm]{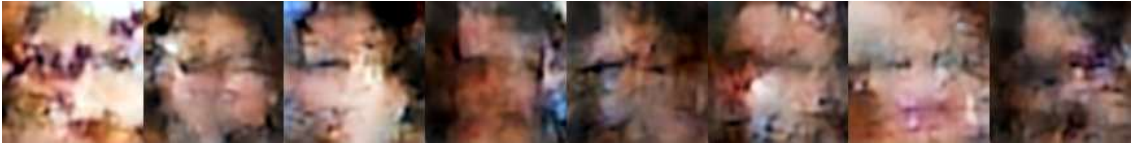}
\end{center}
\label{fig:No_aug.eps}
\caption{Generated images by the trained PC-VQ2 model with no data augmentation.}
\end{figure*}

\section{Related Works}
This section explores the intricate domain of data augmentation, discussing its various aspects and how it intertwines with other data-efficient strategies. 
Furthermore, we clarify the connection between augmentation-based data-efficient methods in generative models, distinctly contrasting them with the method proposed in our research.

\subsection{Data Augmentation in Data-Efficient Algorithms}
This subsection delineates the role of data augmentation within the sphere of data-efficient algorithms. 
It further discusses various concepts and methods pertinent to data augmentation, highlighting the distinctions between phased data augmentation and its counterparts.
 
Data-efficient algorithms have emerged as a significant boon for Deep Learning models, particularly in scenarios constrained by limited data. 
A comprehensive survey by \cite{adadi2021survey} classifies these algorithms into distinct categories: non-supervised paradigms, data augmentation, knowledge sharing, and hybrid systems.
Knowledge sharing includes several strategies. 
These encompass transfer learning, multi-task learning, lifelong learning, and meta-learning.
Our research prioritizes data augmentation. 
This approach is proficient in augmenting images while preserving their labels, often beneficial even when data is scarce within a specific domain.

An extensive review of image data augmentation techniques is presented in \cite{shorten2019survey}. 
The methods are diverse, including kernel filters, geometric transformations, color space transformations, and more.
Our study concentrates on geometric transformations and color space transformations, primarily because these methods maintain the original labels of the data.
When the objective is to augment the volume of images within a targeted domain, kernel filters might not be the optimal choice. 
They are conventionally employed for image enhancement, implying their use should be confined to preprocessing stages where only enhanced images are utilized.

We proceed by contrasting various data augmentation strategies with the phased data augmentation introduced in this paper.
\cite{mikolajczyk2018data} notes that models pre-trained on augmented data and subsequently fine-tuned with original data can learn beneficial information.
However, our proposed method distinguishes itself by altering training data distributions in stages, rather than in a single instance.
Through the gradual limitation or elimination of specific transformations (e.g., rotation), the augmented data distribution incrementally aligns more closely with the original.
Therefore, phased data augmentation can enable models to learn the original distribution more effectively.
Additionally, this approach can enhance the extraction of transformation-related information.

In the context of managing data augmentation parameters, one might consider automated control analogous to AutoAugment \cite{cubuk2018autoaugment}. 
Yet, our method adopts a phased reduction in parameter ranges, acknowledging that automated control typically necessitates additional computational resources and data --- specifically, it demands loss function computation based on validation data.

\subsection{Data Augmentation in Generative Models}
This subsection discusses traditional data-efficient methods rooted in data augmentation within the scope of generative models, providing a comparative analysis with the method proposed in our study.

In the sphere of data-efficient learning, particularly learning with constrained datasets, research focusing on generative models has recently been performed. 
Several studies on GANs \cite{tran2021data,zhao2020differentiable,karras2020training} have independently advocated for the incorporation of data augmentation into the inputs of the discriminator, not only during the optimization of the generator but also throughout the discriminator's optimization process.
This strategy aims to preempt any undesirable shifts in the synthesis distribution and has proven effective for GANs. 
However, it is not applicable to generative models without a discriminator.
For instance, PixelCNNs cannot use this technique.

Another innovative approach detailed in \cite{karras2020training} suggests the implementation of data augmentation under a specific probability, coupled with the aforementioned strategy. 
This technique is further refined by adaptively modulating this probability through a function calculated based on the unique architecture of GANs, reinforcing its specificity to GANs.
Contrary to these methods, our research opts to constrict the data augmentation parameters within certain phases, shunning the use of a gradually diminishing probability. 
This choice stems from the understanding that even a minimal yet non-zero probability could encompass data subjected to the full spectrum of transformation parameters.

A paper on generative models \cite{jun2020distribution} has proposed using data augmentation and conditioning a model with the data augmentation parameters.
However, the study only experimented with a limited array of data augmentation types, for instance, 90-degree rotations, and relied on a more substantial dataset without augmentation compared to the limited data discussed in our paper. 
The task of training a generative model with a mere 100 images necessitates a more diverse set of data augmentation techniques.

\section{Conclusions}
Within the existing landscape of data-efficient training methods, which predominantly cater to GAN architectures, this study represents a pioneering step towards exploring data-efficient training for alternative generative models.  
We have introduced a novel training strategy named ``phased data augmentation," tailored specifically for a likelihood-based generative model operating with limited datasets. 
Our experimental findings consistently demonstrate the efficacy of phased data augmentation, showcasing its evident superiority over the traditional augmentation approach. 
This was validated across various data domains and sampled datasets, indicating consistent performance improvements and robustness for the PC-VQ2 model, even in contexts with limited data resources.
Although we cannot assert that the hyperparameter set used in this study is optimal, it has nevertheless shown consistent and evident effectiveness over the traditional augmentation method across a variety of datasets.
Drawing from the principles of traditional data augmentation, our methodology promises broad applicability to a diverse array of generative models and transfer learning contexts, mirroring the wide-ranging utility of traditional augmentation techniques. 
While theoretically sound, the practical effectiveness of this approach for other likelihood-based models warrants further empirical investigation. 
It is our hope that this research not only advances the capabilities of the PC-VQ2 but also invigorates further study and application of likelihood-based models, facilitating their effective learning from limited data.

\section*{Acknowledgments}
This study was supported by JST SPRING, Grant Number JPMJSP2119.
We would like to thank my supervisor, Associate Professor Kazuhiko SUEHIRO at Hokkaido University for his advice on this paper and review of it throughout this study.
We would like to thank Associate Professor Sho TAKAHASHI at Hokkaido University for his advice.
For enhancing the clarity and accuracy of our paper, as well as for reviewing the code utilized in our evaluation metrics, we employed the assistance of ChatGPT \cite{OpenAI2023ChatGPT}.


\appendix
\setcounter{figure}{0}  
\renewcommand{\thefigure}{A.\arabic{figure}}

\newpage

\begin{figure*}[t] 
\centering
\begin{center}
\includegraphics[width=15cm]{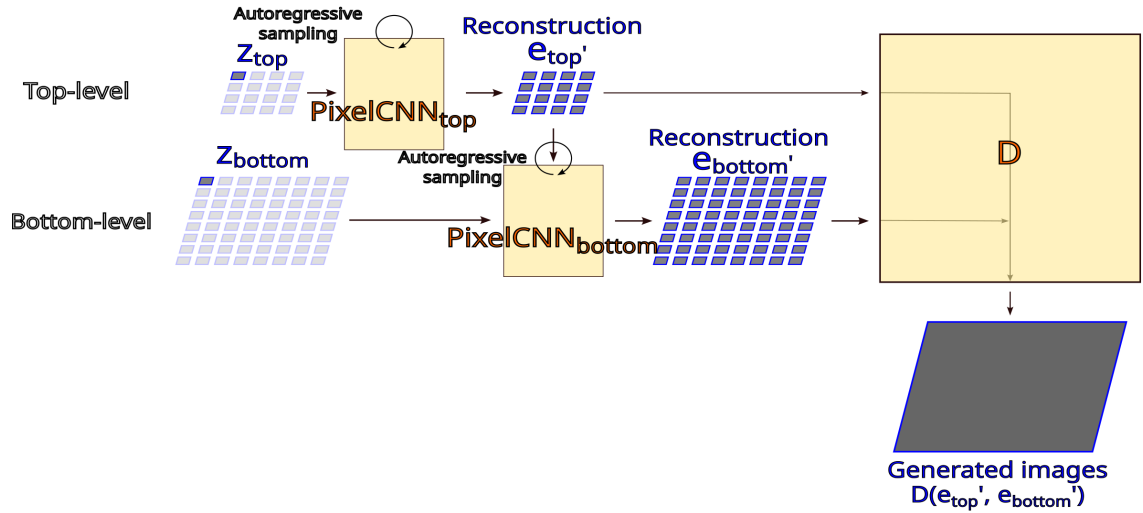}
\end{center}
\label{fig:PC-VQ2_generate.eps}
\caption{Overview of a PC-VQ2 generation structure.}
\end{figure*}

\section{Overview of a PC-VQ2 Generation Structure}
This section overviews a PC-VQ2 generation.
Figure A.1 illustrates the structural layout of the PC-VQ2 generation process.

In each of the discrete latent maps for both the top-level and bottom-level, the value of the top-left pixel represents a discrete vector index, which is randomly sampled from a uniform distribution. 
Subsequent pixels in each map are sampled based on the respective trained PC-VQ models, following a raster scanning pattern.
During the generation process at the bottom-level, the discrete latent maps generated at the top-level are utilized for conditioning, along with the randomly sampled value of the top-left pixel.

\section{Details of the Experimental Settings}
This section details the experiment setups regarding the hyperparameters of PC-VQ2 model and its training.
We implemented FID metrics to evaluate in PC-VQ2, based on the official PyTorch code of StyleGAN3.

We implemented VQ-VAE-2 encoder-decoder, reconstruction, and generation modules on the basis of DeepMind's sonnet library. 
Additionally, we implemented VectorQuantizerEMA of VQ-VAE-2, PixelSNAIL, and GatedPixelCNN, based on Sarus's implementation from tf2-published-models.

For the VQ-VAE-2 hyperparameters, we set the codebook size to 256, representing the number of discrete vectors, and the codebook dimension to 64, indicating the dimension of each vector.
For the PixelCNNs models, we used a dropout rate of 0.2.

Next, we detail the pieces of training.
Our training datasets consisted of images with a resolution of 256 x 256.

Regarding VQ-VAE-2 training, we used Adam optimizer with a learning rate of 0.0003, a batch size of 32, and a total of 4,000 training iterations.

Regarding the PixelCNNs training, basically, we used Adam optimizer with a learning rate of 0.0003, a batch size of 32, and a total of 50,000 training iterations.
Considering that phased data augmentation is akin to transfer learning, the optimizer was reset at each phase transition, and the learning rate was adjusted in subsequent phases, decreased by factors of 10, 40, 100, 500, and 1,000 respectively. 
These factors were empirically determined through trial and error.

To ensure a fair comparison, both the standard data augmentation and the no augmentation experiments employed the same settings.
For standard data augmentation, we changed the parameters of color operations to 0.15.

For a further equitable comparison in the no data augmentation training and when using real images for FID calculations, we employed constant upscaling, mirroring the approach used in phased data augmentation.

\section{Factors Contributing to the Blurry Images Generated by the PC-VQ2 Model.}
This section explains the factors contributing to the blurry images generated by the PC-VQ2 model.

Our model, modified to be more cost-effective than the original, however, likely led to blurriness due to:
\begin{itemize}
\item smaller latent spaces 16-8, compared to the original 64-32,
\item smaller codebook size 256, compared to the original 512,
\item The original FFHQ experiments used a higher resolution of 1024 x 1024 with three latent spaces 128-64-32, unlike our approach.
\end{itemize}

Figure A.2 shows larger latent space configuration, 32-16, can bring out the model's potential to generate sharpened images.

\begin{figure*}[tb] 
\centering
\begin{center}
\includegraphics[width=15cm]{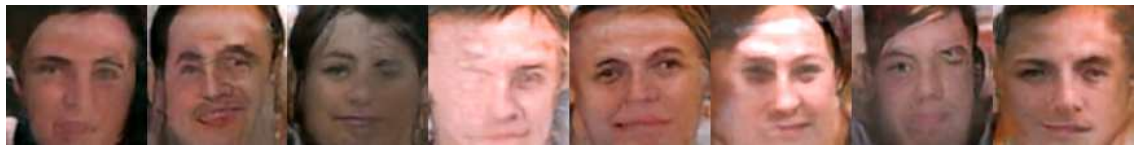}
\end{center}
\label{fig:32_16_phased_75000_best_8_images.eps}
\caption{Human-face images generated by the PC-VQ2 model with higher-resolution latent spaces, 32-16, and phased augmentation.}
\end{figure*}

\end{document}